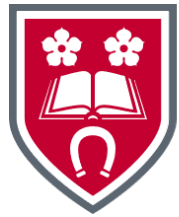

# JobHam- place with smart recommend job and candidate filtering options

Thesis submitted for the degree of
Master of Science
at the University of Leicester

## *By*

## Shiyao Wu

DECLARATION

All sentences or passages quoted in this report, or computer code of any form whatsoever used and/or submitted at any stages, which are taken from other people's work have been specifically acknowledged by clear citation of the source, specifying author, work, date and page(s). Any part of my own written work, or software coding, which is substantially based upon other people's work, is duly accompanied by clear citation of the source, specifying author, work, date and page(s). I understand that failure to do this amounts to plagiarism and will be considered grounds for failure in this module and the degree examination as a whole.

**Name:** Shiyao Wu

**Date:** 09/09/22

# Abstract

Due to the increasing number of graduates, many applicants experience the situation about finding a job, and employers experience difficulty filtering job applicants, which might negatively impact their effectiveness. However, most job-hunting websites lack job recommendation and CV filtering or ranking functionality, which are not integrated into the system. Thus, an smart job hunter combined with the above functionality will be conducted in this project, which contains job recommendations, CV ranking and even a job dashboard for skills and job applicant's functionality. Job recommendation and CV ranking starts from the automatic keyword extraction and end with the Job/CV ranking algorithm. Automatic keyword extraction is implemented by Job2Skill and the CV2Skill model based on Bert. Job2Skill consists of two components, text encoder and Gru-based layers, while CV2Skill is mainly based on Bert and fine-tunes the pre-trained model by the Resume-Entity dataset. Besides, to match skills from CV and job description and rank lists of job and candidates, job/CV ranking algorithms have been provided to compute the occurrence ratio of skill words based on TFIDF score and match ratio of the total skill numbers. Besides, some advanced features have been integrated into the website to improve user experiences, such as the calendar and sweetalert2 plugin. And some basic features to go through job application processes, such as job application tracking and interview arrangement. This paper has been divided into seven parts. The first part introduces the aims, objectives and challenges of the project. The second part provided this project's three requirements: essential, recommended and optional. The third part introduces the background of the project and some related techniques that will be used in the project. The fourth part describes the project's system design, including the frontend, backend, model architecture and ranking algorithm. The fifth part shows the details of implementing the functionality and automatic keyword extraction. The sixth part describes the evaluation methods for the website, and the last part summarises the project and future work.

# Table of Contents

## *1. Introduction*

This section contains aims and objectives for the project and present the existing challenge.

### 1.1 Aims and Objectives

The aim of this project is to develop a job hunter web application which can help graduates and head hunt find suitable position and potential candidate separately. According to the study, it shows that the number of graduates take a slight upward trend recently [1]. Besides, the number of undergraduates climbed from 492,355 in 2019/20 to 527,070 in 2020/21, which increased by 7%; similar to undergraduate, postgraduates' figure also growth by 12% in 2020/21 [1]. However, over 96,000 recent graduates (those who graduated in 2020 and later) experience unemployment at the current rate of 12%, which means that around 12% graduates cannot find their job when they leave higher education [1]. Therefore, it is important for them to find a suitable job after graduation, which can be competent with their skill in a short time, especially for those who do not have relevant experience of job. Besides, faced with an increasing number of graduates, head hunt will receive many CV, which is a more challenging task for them. According to research, processing un-structed datasets, particularly for resume and CV data, is more difficult and can cost employees up to 80% of their daily revenue [1, 2]. Therefore, how to develop a cost and time effective job hunter website is a key question.

### 1.2 Challenges

When processing a large number of CVs, it is important to extract basic and essential information from PDF format files. At the same time, extracted skills from the job description would be matched with important details from the CV. While training with the job description to skill machine learning model, the job description dataset would be gained from Kaggle. Besides, data pre-processing with a raw job description dataset would be the first step of the keywords extraction task. Once get the extracted skill from the job description, the calculation between extracted skill and importation information from the CV would be essential to implementing job recommendation and CV filtering. This challenges keyword extraction and matching matric using natural language processing based on a machine learning model. While conducting the project, some advanced features would be integrated into the website, such as a calendar, notification centre, etc. Therefore, integrating the above functions would be another challenge in this project.

## *2. Requirements*

This section contains essential requirements, recommended requirements and optional requirements related to the project.

### 2.1 Essential Requirements

In this application, users can be separated into three accounts, including the head hunt, job applicant and admin. Except for admin, users can register and log in to the account by filling in their basic information. A job applicant can search job by its title, company, location and type (i.e. full-time or part-time) related to the job; they can upload their CV and preview their CV in pdf format; they can track their job application status once they apply for a job and receive an updated notification with important information once they saved their preferred jobs (i.e. job application deadline). Head hunt can upload job advertisements and have a list of job applicants for each job. Besides, the admin can delete the head hunt and job applicant accounts.

## 2.2 Recommended Requirements

The integrated calendar on the website records important dates, including job application deadlines and interview dates. A job dashboard would be integrated for the head hunt to have an overview of data visualization, including the high related skill from the job description, Word cloud from the job description and ranking candidates with their match ratio with the job description. Interview arrangements would be provided with available time slots from the head hunter based on the integrated calendar.

## 2.3 Optional Requirements

Automatic Keyword extraction from CV and job description would be implemented to obtain basic information from CV and skills from both. Skills matching algorithm would be an important way to implement CV filtering and job recommendation based on TFIDF. For job applicants, recommended job list would be provided on the website based on their CV. For the head hunt, job applicants would be ranked and provided on the website, which is helpful to get shortlist applicants.

## *3. Background Research*

This section introduces the technology, framework and recommendation approaches applied in the project, including MVC, Vue.js and so on.

## 3.1 MVC Architecture

Multiple views are presented based on the MVC Design Pattern, and it has been the most popular web application architecture. It divides the web application into three parts: Model, View, and Control. View represents the website's user interface, including HTML and CSS; controller controls user logic, call API and receive data from the model; model provides API and process transferred data. By separating the business logic from the user interface using MVC, it is possible to create applications that are simpler to manage and maintain since both designers and programmers can alter the underlying business rules without negatively impacting the application's visual appearance [3].

## 3.2 Vue.js

Vue.js has been known as a frontend development framework for user interface, which works on top of HTML, CSS, and JavaScript. Programmers can quickly develop user interfaces as its declarative and component-based programming methodology. In this framework, single file components (SFC) would be provided, which contain the component's logic (JavaScript), template (HTML), and styling (CSS) in a single file [4]. Besides, a programmer could also separate HTML, CSS, and JavaScript into three files. Therefore, Vue.js could help programmers develop and maintain the frontend more easily.

Besides, Vue.js contains many toolkit libraries, which could be applied in the frontend and improve user experience. For example, while dealing with the integrated calendar in this project, V- Calendar library would be applied and shown on the website. In this library, programmers could customize the calendar and mark the important date on the calendar [5]. And Bootstrap library could provide functions on website layout, which is helpful to conduct frontend.

## 3.3 Flask

Flask is a Python framework for building the backend, and it is a micro-framework to construct web applications quickly. Compared with Django, Flask performs faster generally and is lightweight. The complete request object, endpoint routing system, etc. are some features offered by Flask [6]. The above essential functions are provided by Flask, allowing programmers to add features as needed during implementation.

## 3.4 Firebase

Web application database platforms like Firebase are popular. It provides many services to handle data and API to operate data. Firebase, the system's backend, functions as a database to store data. It utilizes JavaScript Object Notation (JSON) standard to store the data and does not require a query to add, delete, or update data, allowing developers to build high-quality applications, including IOS, Android and Web [7]. This platform provides Firebase Auth, Firebase Storage and FireStore Database. Firebase Auth supports user authentication with email, password, and other social accounts, such as Google, Facebook, etc. User management is offered in this service. Firebase Storage, cloud storage to store a file, could be used to store images, audio, and other customized content [7]. FireStore Database performs as a database to store user data and static data related to the website, and it offers essential operation with the database, allowing developers to complete the database in a minute.

## 3.5 Automatic Keywords Extraction

The method of selecting words and phrases from a written document that, based on the model and without human intervention, can best reflect the essential and significant contents is known as automatic keyword extraction [8]. Even though it has been used in many applications, such as document summarization, it might be challenging to extract keywords from a broad text since we need to know how they connect to the paragraph [9]. In the domain of keyword extraction, there exist four methodologies, including simple statistical, linguistics, machine learning, and hybrid approaches [10].

Regarding simple statistical approaches, techniques, such as term frequency (TF) and term frequency-inverse document frequency (TF-IDF), basically, calculate the frequency of occurrence of the word to determine whether it is a keyword. Similarly, word co-occurrence algorithms can be represented by the frequency of a word and the frequency of its occurrence with other words, which is then computed confidence of the words using this statistical data. Compared with a simple statistical approach, the linguistics approach makes use of the linguistic characteristics of the words to identify and extract keywords from text sources [10]. However, the above methods are mainly based on statistical analysis without training datasets and are not recognized as a learning problem. Machine learning provides a way for keyword extraction to help the machine learning model understand the meaning of related paraphs, also known as natural language processing. This method requires annotated training dataset to assist the model in understanding language and obtain a state-of-the-art performance compared with the above methods. Hybrid approaches integrate the above methods with word features, including position, length and so on.

Considering the performance of the methodology, machine learning would be the preferred method in this project. Bidirectional Encoder Representations from Transformers, known as Bert, would be the backbone in many natural language processing tasks since it has been pre-trained with two basic natural language processing tasks: Masked Language Model (LM) and Next Sentence Prediction (NSP). And it aims to understand the relationship between sentences [11]. Therefore, it could be better for other natural language processing tasks, such as keyword extraction.

## 3.6 Job Recommendation

Widely used in many applications, recommender systems recommend goods, services, and information to potential clients. According to current research, there are more job advertisements for job applicants generally. Hence many e-commerce platforms offer job recommendation services to shorten the time spent hunting. Job recommendation means that the top n jobs matched the application profiles for jobs. To make job recommendations for each user cluster, three recommendation techniques would be mainly applied: Content-Based Approach, Collaborative Filtering Approach and Linear Hybrid. The content-Based approach will convert content into n-dimensional features. Each dimensional feature represents a word in the sentence, and the TFIDF score determines the value of the dimensional features. While Collaborative Filtering will predict the implicit features based on the provided features, for example, if there exist some job applicants to the specific posted jobs, this method will also identify similar users compared with applied job applicants by calculating the similarity between users. It will also recommend the job to the user if the similarity is more significant than a threshold. Besides, as the features differ for each user, user clustering based on K-means has been applied in the system iHR [12]. The linear hybrid method combines the content-based approach and collaborative filtering, which take advantage of each side. It provides n-dimensional features to represent the content of the job and calculates the similarity between job applicants and users. This project has applied the content-based method to provide n-dimensional features for each word, but filter skills feature in the total word list.

## 3.7 CV Ranking/Filtering

Many IT companies usually get many CVs for a single job posting; hence they encounter a similar information overload issue. The recruiters had to review each application and CV, which was tiresome and time-consuming. Therefore, these businesses attempted to apply technologies to help in the resume screening process for this reason, which could improve effectiveness during human recruiting. Applicants ranking means that the top n job candidates would be the ranking item for job applicant ranking/filtering. Traditionally, it would provide a job profile based on a job advertisement, and list required skills in the job profile and user profile, then matching candidates and jobs would be based on listed skills and provide a ranking list [13].

Latent Semantic Analysis (LSA) was applied by Lu et al. [14] to determine the similarities between jobs and applicants, although they only compared candidates based on two factors: interest and education. An expert method was provided by Drigas et al. [15] to match positions with job seekers and suggest unemployed people with the job vacancy. The system calculated the similarity between user profiles and job postings based on Neuro-Fuzzy, which requires training dataset to train Neuro-Fuzzy network. A hybrid recommender system that combined content and interaction-based relationships was also presented by Yao et al. [16]. The similarity of their profiles allows the content-based section to identify relationships between job-job, job-job applicants, and job applicants-job applicants. When dealing with unstructured data, the similarity between a job and profile based on latent semantic analysis is applied. Weight sum values are returned for structured data, such as age and gender.

However, those methods mainly focus on the user data instead of the applicant's CV, as many experiences and skills would be mentioned in the CV, and human resource management could glance at the CV and select the applicant, especially in the formal process of human recruiting. Therefore, information about job applicants would be better from a CV instead of common user data stored on the website.

## *4. System Design*

This section introduces the system architecture, frontend, backend, and overview of Job to Skill model and its related training dataset.

### 4.1 System Architecture

This system applied CSS HTML layout and was built on top of existing UI templates (The Hunt by UIDeck Company [17], while the underlying data structures and backend were developed from scratch based on the vue.js framework. The backend for this system can be separated into two parts, including Flask to provide REST API and a machine learning model to implement job recommendation and CV filtering.

### 4.2 Frontend

Based on the existing UI, the user interface would be provided in the frontend. Besides, user logic input and database essential operations related to the Firestore Database would be controlled by JavaScript. For example, user log-in and register operation functions are implemented in the frontend instead of the backend.

There exist three website roles: admin, hunt and applicant. Admin can delete the account of hunters and applicants. Hunt can post a job with the required information, such as job title, job company, location and etc. Hunt can manage jobs, including deleting and updating and controlling the job application process, arranging interviews, providing job offers, or rejecting job applicants. What is more, to help the hunt deal with the job application more effectively, ranking job applicants have been provided on the website; on this website; the hunt can arrange interview based on the ranking of job applicants, which definitely help the hunt improve their effectivity while hiring the employees. Also, they can check the job dashboard related to the job application and skills related to the job, which includes scored skill extraction and ranking job applicants with their scores. Job applicants can process basic job application steps, such as applying for the job, saving job details and checking the job application status. To help employers get high-scored applicants, the applicant will upload their CVs. Based on their CV, they even can receive a ranking list of posted jobs, which can help applicants to get their jobs in a minute.

To improve user experience, some advanced features would be integrated into the frontend based on some libraries. For example, integrated calendar is based on toastUI calendar library [18]; head-hunter dashboard with job to implement data visualization is based on Echart [19]; PDF previewer is based on VuePdfEmbed [20]; GET request from frontend in JavaScript to backend is based on axios tool [21].

### 4.3 Backend

The backend provides related API for frontend, such as JobMatchCVAPI, CVMatchJobAPI, WordCloudAPI, Job2SkillAPI. JobMatchCVAPI could provide a list of ranking CVs to match job description obtained from job ID in Firebase. CVMatchJobAPI provides a list of ranking job to match applicant CV obtained from job applicant ID. WordCloudAPI could generate word cloud picture and upload to firebase. To implement data visualization with extracted skill from job description, Job2SkillAPI provides related skills from job description. Besides, CV and job description matching algorithm is based on TFIDF and match ratio. TFIDF stand for the frequency of skill in the job description sentences and match ratio stand for the ratio of matched skills to total skills in job description. Automatic keyword extraction processing to parse job description and CV is based on Bert. Extracted skills from job description is based on Bert Multi Label Text Classification model [22]. Extracted skills from CV is based on Bert for Token classification model [23], which is mainly used in name entity recognition, though some related information would be extracted as well, such as designation, college name and years of experience.

The database is stored in Firebase. Three collections exist, including 'User', 'Job' and 'Application'. Even though three roles are stored in the same collection, 'User', the data is different from each other. Except for particular information, all users will store name, email, role and uid. For the hunt, the data also contains the job_list to include a list of job IDs they posted on the website. For the applicant, the data also contains the apply_list and saved to store a list of applied job IDs and saved job IDs. For collection 'job', it will have job id, job company, job title, description, salary, type, and interview list of applicants. For collection 'Application', it will store the list of job applicant IDs in each job id document, which is easy to be accessed by the hunt. Besides, it will keep the applicant's CV and WordCloud for the job description. In the Firebase Storage, two folders, CV and wordcloud, exist to store related files. Also, the user authentication information is stored in Firebase Authentication; uid, email address and password are stored in it.

## 4.4 Job2Skill model

The job to skill model is inspired by Bert Multi Label Text Classification model [22] and GRUBERT [24]. In this model, Text Encoder [23] and Gru-based Layer are two major components, which is shown in Figure 4.1:

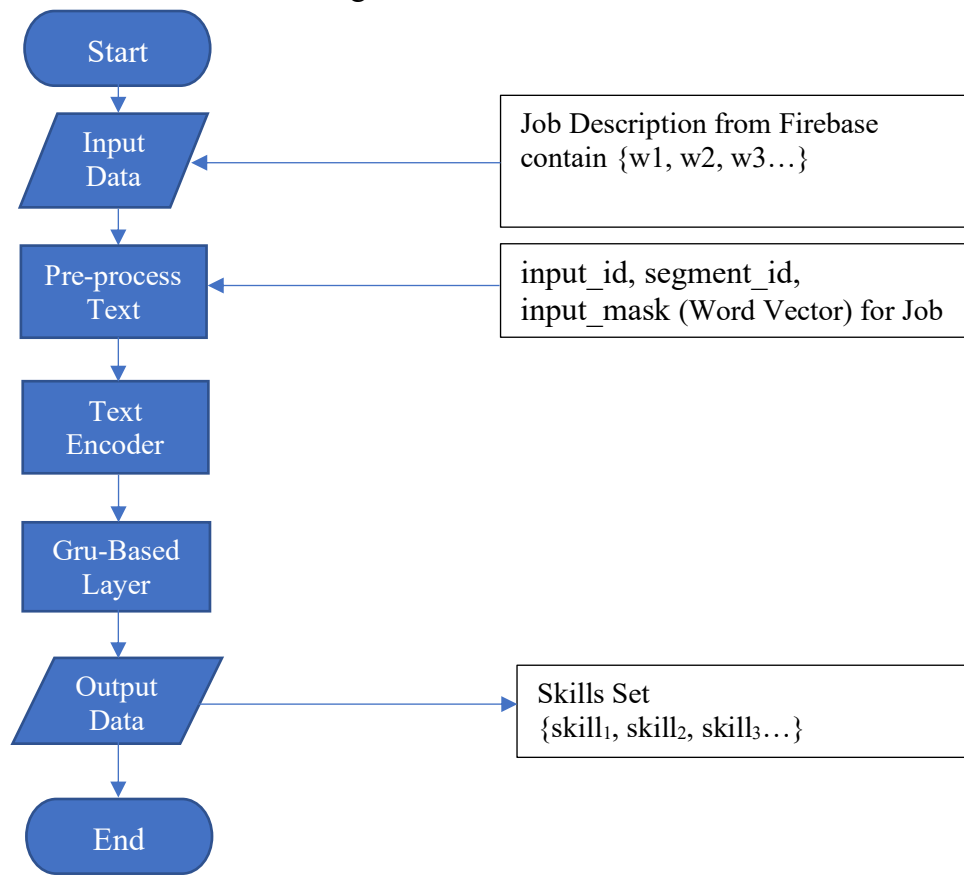

Figure 4.1 Job2Skill Model Architecure

**Pre-process Text:** The figure shows that the input data is a job description from Firebase, which contains several words (w1, w2, w3, and etc...). The pre-process text component will change the sentence to the token by BertTokenizer with Bert vocabulary in the pre-trained model 'base-uncased' and add [CLS] and [SEP] token, which represents the start of the sentence and the end of the sentence. As The sentence is the first sentence of the document, the segment id will be marked with 0 with the length of the token. BertTokenizer will convert its token to the token's id based on the WordPiece algorithm, also called the input_id vector. If there exists input_id, the input_mask vector will fill it with 1. The vector length should be the same when fed into the Bert model. Therefore, the remaining vector will be set to padding, which is set to 0. After pre-processing the text component, input_id, segment_id, and input_mask will be generated at this stage. As the max length of the sentence is imposed to 256, the shapes of input_id, segment_id and input_mask are epoch*max_length, which is equal to [1*256].

**Text Encoder**: The vectors with input_id, segment_id and input_mask are the input data for text encoder. At this stage, the pretrained model will learn what is language and what is context using Masked Language Model (MLM) and Next Sentence Prediction (NSP). As the pretrained model 'Bert_base_uncased' has 12 layers of Transformers block, 768 hidden layers and 12 self-attention heads, the shape of output from BertModel is [1*768]. And Fine-tuning will be processed with its specific tasks on training data and its label at this stage. The dataset for training is obtained from GitHub [25], which is from Singaporean government website and contain 20, 298 structured job posts and has 2,548 distinct skills. After this stage, the trained model can deal with the skill extraction from job description tasks.

**Gru-based Layer**: Inspired by GRUBERT [24], text Encoder will follow the Gru-based layer. Obtained by text encoder, embedded text representation with pooled_output will be fed into a Gru layer. This layer will extract important features from low-level representation by utilizing gated recurrent units (GRU) as the gate in the Gru will learn the information if it should be filtered or not. And the shape at this layer will not change, still with [1*768]. The representation would be normalized between -1 and 1 based on the tanh function and get the output to Dense Layer. The dense layer contains two linear layers and the Relu function, which are used to downstream the previous result to the number of labels and alleviate the situation of the Vanishing Gradient Problem separately. The first linear layer will be fed with the output of the Gru layer, which is input with the [1*768] matrix and output with the [1*2000] matrix. As the number of labels is 2548, the second linear layer will be fed with a matrix after the Relu function and has an output with the [1*2548] matrix. Relu function is a common function used in many deep learning models. It will remain the original input if the input is larger than 0; otherwise, the output will remain as 0.

### 4.5 CV2Skill model

The CV to skill model is a fine-tuned name entity recognition model based on Resume-NER [25]. In this model, data tokenization, pre-training and fine-tuning are significant steps.

**Data tokenization**: It concludes by adding tokens and special tokens, such as [PAD], [SEP] and [CLS]. Token [PAD] is used to make every sentence have the same length. BertTokenizerFast from the transformer could generate input_id and attention_mask features, which will be fed into the model. The shape of input_id and attention_mask is [1*500] as the max length is set to 500. Therefore, if the sentence is longer than 500 words, it will be cut off the extra words; otherwise, it will fill with the [PAD] token.

**Pre-trained model & Fine Tuning**: 'Bert-base-uncased' will be used as architecture, which is the same as the Job2Skill model. The training dataset and testing dataset 'Resume Entities for NER' were collected from Kaggle [26], and it is split into 180 data for training and 40 data for validating. The training data will be fed with an entity label. A list of an entity contains name, Designation, College Name, Years of working experience and skills. Therefore, a model for the Resume-Name Entity Recognition task has been trained.

## 4.6 Job/CV ranking algorithm

Skill matching algorithm computes match of CV and job description based on scored skill. The scored skill is provided by the TFIDF algorithm, which stands for Term Frequency (TF) and Inverse Document Frequency (IDF) separately. IDF calculates the percentage of documents that include the words, whereas TF computes the number of times a word appears in a document. TFIDF will combine IDF and TF by multiplying these two scores. For example, if the word 'dog' has occurred 20 times in a 100 word-document. Therefore, the score of TF is 20/100, equal to 0.2. The size of the provided corpus is 10,0000 documents, and the word 'dog' has occurred in 20 documents. Then the score of IDF is log (20/10,0000). Therefore, the score of TFIDF is the result of IDF multiplying TF, that is, 0.2*log (20/10,0000). Even though TFIDF will compute all words in the sentence, extracted skills from CV and job description will help filter words with related skills, which would stand for the importance of skills. The match ratio between CV and job description will be calculated as the proportion of matched skills in the total number of skills in CV or job description, which is shown in Formula. Therefore, the final score will be provided by multiplying scored skills and match ratio, which is shown below:

$$match_{ratio} = \frac{len(matched_{skill})}{len(total\_num_{skill\ from\ job})} * 100\%$$

$$Scores = scored_{skill} * match\_ratio$$

## *5. Implementation*

### 5.1 GUI

GUI has been implemented using the Vue framework and JavaScript, and it has been split into different components based on other users, including Admin, Applicant, Hunt, and an unregistered user.

For unregistered users, they can register a user account, fill in basic information, and choose different roles with their account, such as Applicant and Hunt, which stand for the person who is looking for the job and is hiring employees separately. The required information for this component is the full name, email address, password, confirmed password and role, as shown in Figure 5.1. If any of the required information is empty or the password does not match in two typing inputs, it will pop up an alert to the user. Besides, the registration process will fail if the email has been registered. Also, the length of the password should be at least six words. Once the user clicks on the Register button and all input information are satisfied the requirement, it will automatically sign up with its email using Firebase Authentication API, store user information and initialize data in the collection 'User' in Firebase Cloud Database. The user data includes uid obtained from Firebase Authentication and required information by user input. Some initialized data will conclude CV_path, apply_list, and saved list for applicant and job_list for hunt.

Figure 5.1 Registration Page

After a user signs up for their account successfully, they can log in to their account using their email address and password. For the different roles of the user, the navigation will show separate components. For admin, the navigation will show the job page and manage an account in Figure 5.2. For user who does not log in, it will show the job page and sign-in page in Figure. For the applicant, it will show the job page manage account and cv in Figure 5.3. For the hunt, it will show the job page, manage account, and post job page in Figure 5.4. Besides, the managed account for applicant, hunt and admin is not the same. The manage account page will show manage jobs, job application, calendar, and notification page for the hunt, which is shown in Figure 5.6. Compared with the hunt, the manage account page will show the resume, saved jobs and job recommendations instead of the manage job page and job application page, which is shown in Figure 5.5. For admin, the manage account will show the manage hunt and applicant page shown in Figure 5.7.

Figure 5.2 Navigation for admin

Figure 5.3 Navigation for applicant

Figure 5.4 Navigation for Hunt

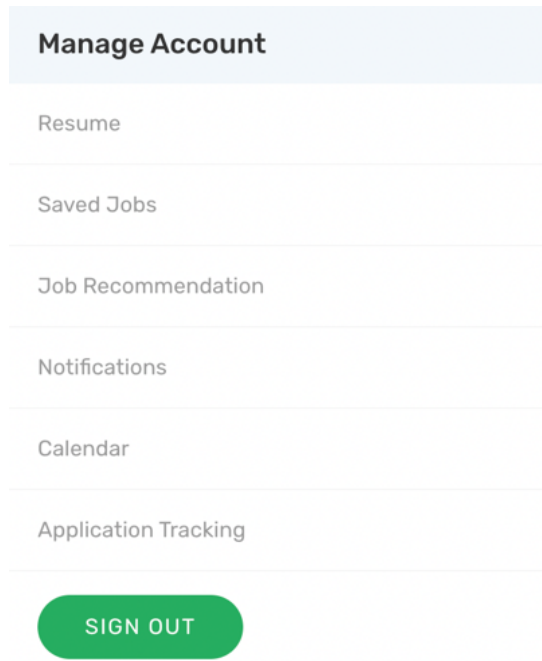

Figure 5.5 Manage account for Applicant

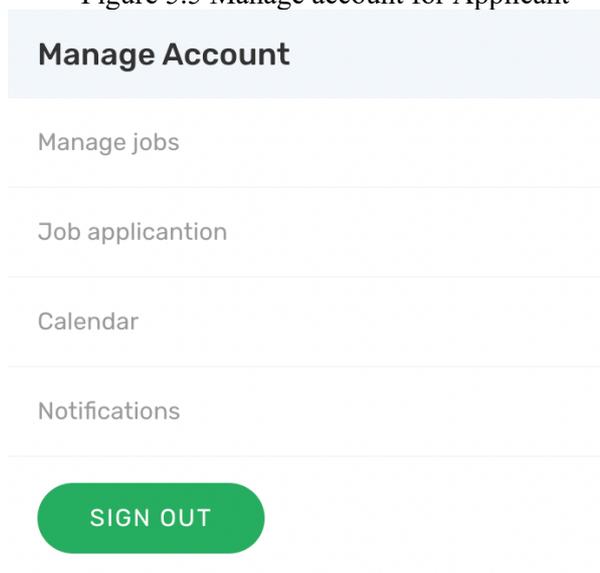

Figure 5.6 Manage Account for Hunt

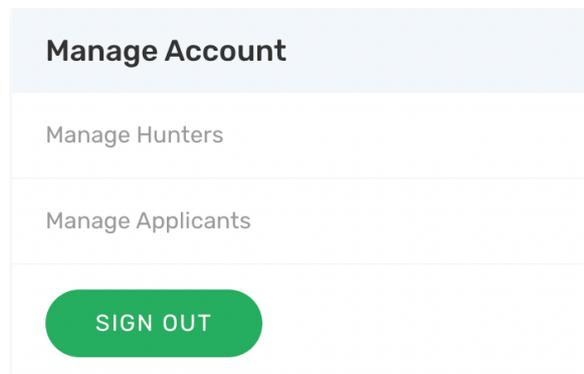

Figure 5.7 Manage Account for Admin

For all users, they can browse the job page and detailed information on the job page, and the following page is the job page for all users shown in Figure 5.8. On this page, they can search for jobs with job titles, company names, job types and locations. Once the user inputs the search content, it will look through the collection 'Job' database and return the specific research result to the user based on the search content. The detailed information job page will show the requirements or description for the job, and other detailed information, including annual salary, the application deadline, location etc. The applicant can click on the save job and apply job button to record their application information. Otherwise, it will pop up an alert about the authentication denied information.

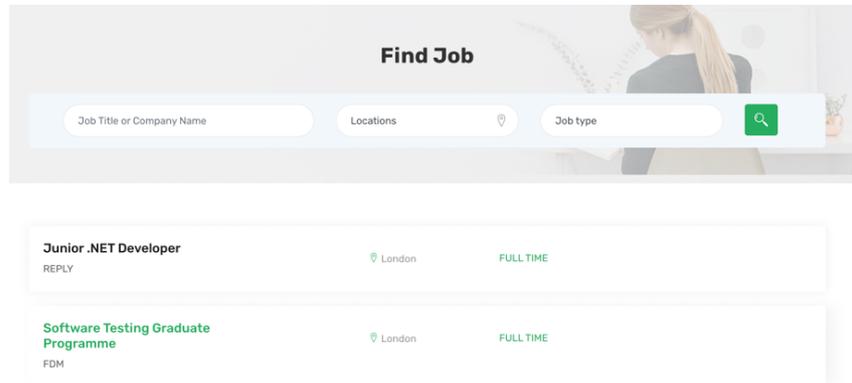

Figure 5.8 Job Page for all user

The applicant can access the resume page on the manage account page shown in Figure 5.9. On this page, users can upload their CV or preview their CV in pdf format after uploading it. When the user clicks the upload CV button, the user can choose the CV file from a laptop, which will be uploaded to Firebase Storage and saved CV path to the user database. As the automatic keyword extraction from CV has the file format limitation, it will pop up an alert when the user does not choose PDF format. Once user upload their CV successfully, they can preview their CV on the same page based on VuePDFEmbed [20]. At this stage, the download URL will be accessed by the CV path stored in the collection 'User' database and VuePDFEmbed will show a preview of the PDF on the website with page operation.

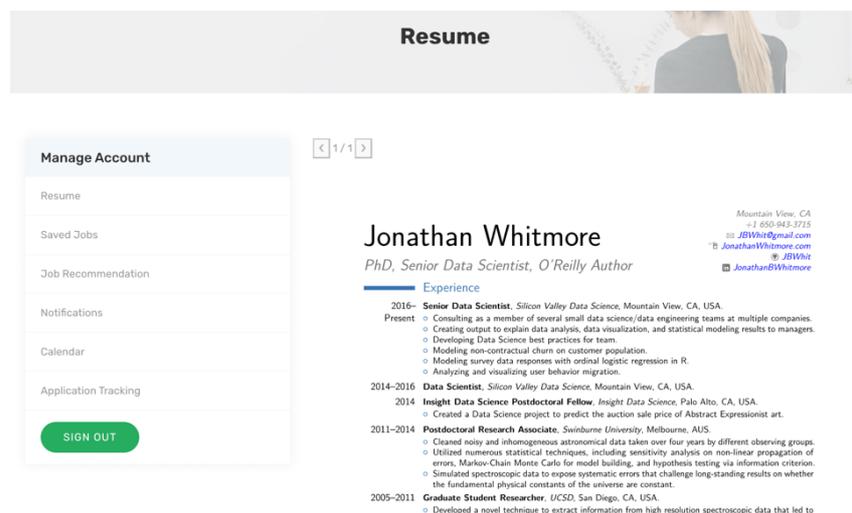

Figure 5.9 Resume Page for Applicant

The saved job page will show the saved job list for the applicant if the applicant would like to receive notification about this job, including the job application deadline, shown in Figure 5.10. And applicant can also delete their saved job on this page. Once the user clicks the delete button on the page, the saved job list will be deleted this job information and updated saved job list on Firebase collection 'User'.

Figure 5.10 Saved Page for Applicant

The notification page will show the application deadline and interview date of the applied job, only the application deadline of the saved job for the applicant, and only the interview date for the hunt shown in Figure 5.11. On this page, if a user is an applicant, it will collect data for the applied job list and saved job list from collection 'User', which contains the job id in those lists and interview data from collection 'Job', which contains applicant id and interview date time. If the user is the hunt, it will only collect interview data from the posted job from the user.

Figure 5.11 Notification Page for Applicant

For the applicant, the calendar page will show a specific date for the application deadline and interview time, as shown in Figure 5.12. The data collected from Firebase is almost the same as the notification page. For the hunt, it will show a specific interview date and time and arrange an interview with the applicant. Regarding interview arrangements, the hunt can select date and time on the integrated calendar. Once they click on the 'saved' button, the interview data will be stored in the Firebase collection 'Job', which includes the applicant ID and interview date and time. After the interview arrangement has been completed, the updated interview will show on the calendar.

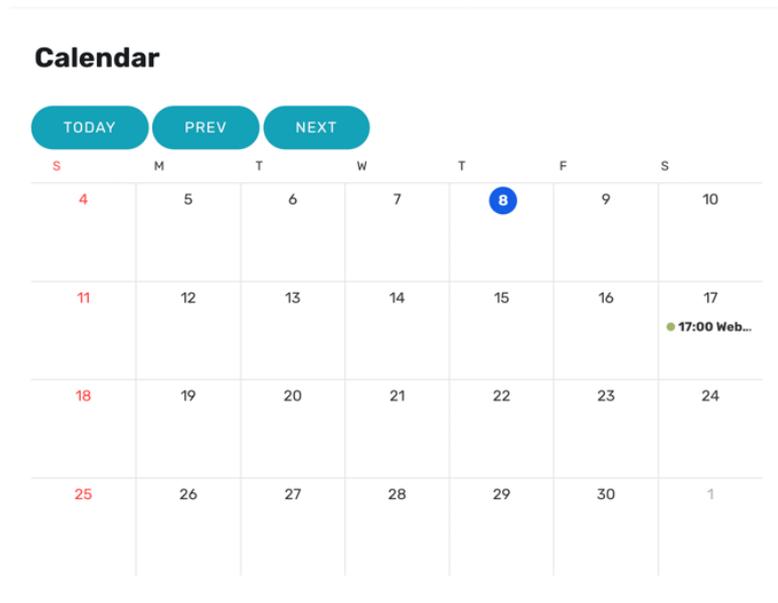

Figure 5.12 Calendar Page for Applicant

Applicants can track their job application on the Application Tracking page, shown in Figure 5.13. On this page, the detailed information for job application will be shown on the page, including application statuses, such as offer, interview process, and rejection, which is stored in the Firebase.

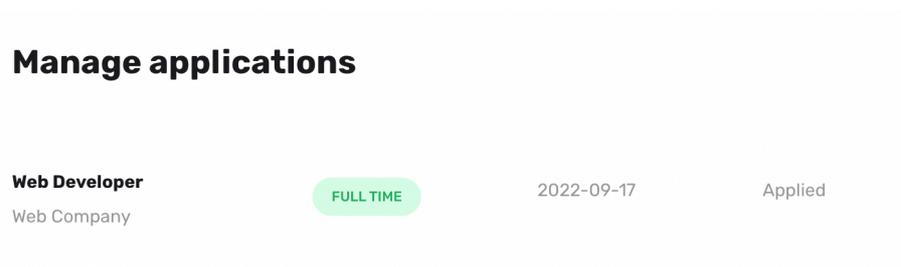

Figure 5.13 Application Tracking Page for Applicant

For admin, they can manage applicant or hunt accounts, such as delete operation, which is shown in Figure 5.14. On this page, if the admin clicks on the delete button, the account will be deleted from the Firebase collection 'User' and authentication record.

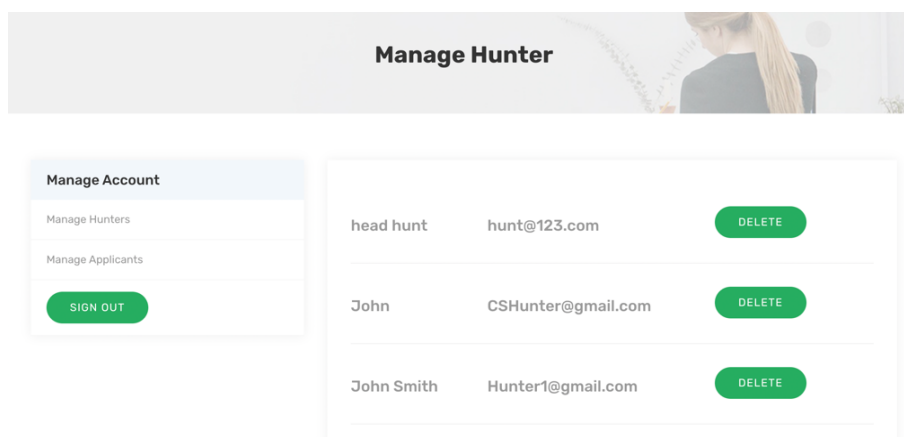

Figure 5.14 Manage Account Page for Admin

For the hunt account, they can post a job by the navigation button, shown in Figure 5.15. On this page, the user will input some required information, such as title, location, salary, type, company, job description and application deadline. Once the user inputs the required information, the new document with the posted job will be added to the Firebase. Besides, the latest job id will be added to the record of job_list in the hunt account.

Figure 5.15 Post Job Page for Hunt

The manage jobs page only can be accessed by the hunt account, and they can manage posted jobs, which is shown in Figure 5.17. The list of the posted job by the specific user will be shown on this page, and the operation contains update operations and deletes operations. Once the user clicks the delete button, the posted job document indexed by its id will be deleted in the collection 'Job'. Besides, the user will jump into the update page with the specific update job page after they click on the 'update' button, which is shown in Figure 5.16. On the update page, the user can preview the current information about the job. When they would like to update some information about the job, they change the value shown on the website. The updated information will change in the specific job document.

Figure 5.16 update Job Page for Hunt.

Figure 5.17 manage Job Page for Hunt

On the job recommendation page, the applicant can receive a list of ranking jobs on this page. Once the user enters this page, the user id and apply_list will be passed to the CVMatchJobAPI, and the returned list of the scored job will be shown on the page shown in Figure 5.18. Based on the scored job list, a ranking list of the job will be provided, and the basic information about the job will be obtained from Firebase and shown on the page.

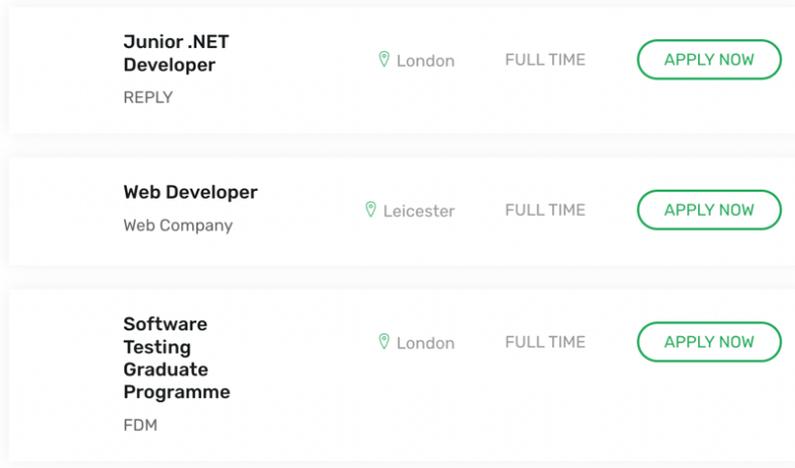

Figure 5.18 Job Recommendation Page for Applicant

In the job application for the head hunt, they will have a list of the job they posted, shown in Figure 5.19. For the operation with the job, they can either choose to check applicants or the job dashboard. If they click check applicants, a list of ranking applicants will be shown on the page, which JobMatchCVAPI obtains. Once the user enters the check applicants page, the specific job id and job applicants who have applied to this job will be passed to JobMatchCVAPI. A list of ranking applicants will be shown on the page. Besides, the matched skill with the job description will also be shown after the applicant's name. If the user clicks on the job dashboard, it will show three charts, including wordcloud for the job description, scored skill extraction for job description and scored job applicants.

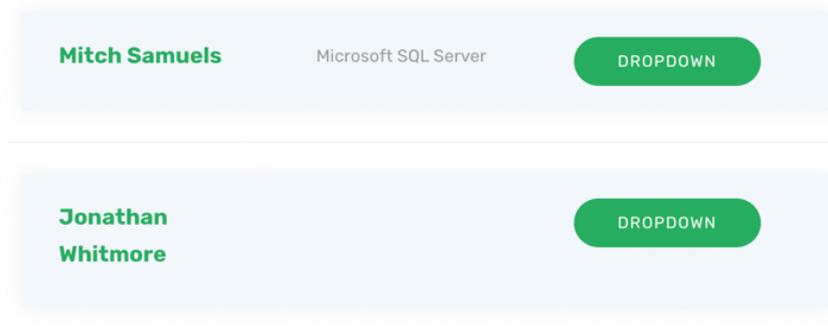

Figure 5.19 Manage Job Applicant Page for Hunt

## 5.2 Automatic Keyword Extraction

In this part, keyword extraction from CV and job description will be implemented based on Bert and provide test code to extract skills with a trained model.

### 5.2.1 Information Extraction from CV

As the limitation of Resume Name Entity Recognition model training on my computer, the model training performs online on a third-party platform, AutoDL [27] and the trained model will be loaded on my laptop. All the information here would be extracted from the backend. In the backend, the CV root will be provided by the user and the extracted information would be the results and stored in the Resume_Info class. Accessing this class, basic information can be obtained by its features, such as Name, YearsExperience and Skill feature in Resume_Info class. Entity id will be obtained by filtering unwanted unique tokens and selecting the entity id with its max value in the output data, which means the selected entity id has the highest probability compared with other entity labels. If the extracted skills from the CV matched with extracted skills from the job description, Skill features from Resume_Info would be returned. The test codes provide a way to extract skills from a CV in a minute. It first gets basic tokenization based on BertTokenizerFast and load vocabulary. After loading the trained model, it will pre-process CV, which contains the removal of punctuation. The processed CV text and entity id will be fed into predict function. In this function, the input id and input_mask are generated from tokenizer.encode_plus with its padding to fill in the length to its max length. The returned result is the probability of each entity id, and it will select the highest chance of entity id, which can be changed to a specific entity name.

### 5.2.2 Skill Extraction from Job Description

For skill extraction from job description, the model train on the AutoDL platform [27] and it will be loaded on my laptop. Word vector would be obtained from this component by converting text into low-dimensional dense vectors based on pre-trained Bert model [23]. As it is considered as classification task in Bert, token [CLS] would be added to the start of the sentence and token [SEP] would be added to the end of the sentence. Generated input id is obtained by looking up words in the pre-trained Bert model and replacing it with sub-word and specific token, which is based on WordPiece algorithm. The processed input sentence, generated input id, will be fed into Bert model. A list of skills id related to the job description will be provided and remain the first 20$^{th}$ skills id, which will be changed into skills label.

### 5.3 CV Ranking

Once Hunters log in to the website, if the posted jobs have applicants, they can preview the candidate ranking results on the check_applicants page. On this page, the candidates will be ranked based on their match ratio and show matched skills with the job description. Once hunters enter this page, the job id and applicants' id to this job will be transferred to the backend by JobMatchCVAPI. Once it gets a list of applicants' IDs and job IDs obtained from the frontend, it will get a list of applicants' CV paths and pre-processed job descriptions by removing punctuation. And the sequence of the CV path will be thrown into the Advanced_JobMatchCV function with the job description. Advanced_JobMatchCV function will firstly extract skills from CV and job description. The TFIDF score will be provided with extracted information from the job description, which is implemented in Code snippet 5.20. However, the list of extracted information based on TFIDF contains many words which are not related to the skills. Therefore, the operation to filter skill words is an essential step. A scored list of skills could be obtained after the filtering operation, which is implemented in Code snippet 5.21. The skills list extracted from the CV will match with the scored list of skills and remain on the matches list. The match ratio will be calculated between the length of the matched list and the length of the skill list for the job description. Thus, returned value contains the match ratio, matched skill list and applicants' names. Therefore, the return value will be the ranking applicant based on the match ratio, which will be shown on the browse_resume page.

```python
# get tfidf vectorizer toolkit
tfidf_vectorizer = TfidfVectorizer(use_idf=True)

job_vectors = tfidf_vectorizer.fit_transform(JobDescription)

# place tf-idf values in a pandas data frame
df = pd.DataFrame(job_vectors.T.todense(), index=tfidf_vectorizer.get_feature_names(),
                  columns=["tfidf"])
result = df.sort_values(by=["tfidf"], ascending=False)
```
Code snippet 5.20 Scored Keyword Extraction based on TFIDF

```python
for re_index in result.index:
    for job_skill in Jobskills:
        m = re.search(re_index, job_skill, re.IGNORECASE)
        if(m):
            ratio = len(re_index) / len(job_skill)
            score = result.loc[re_index]['tfidf']
            scored = {
                'skill': job_skill,
                'score': score,
                'ratio': ratio,
                'match': re_index
            }
            scored_index.append(job_skill)
            scored_result.append(scored)

scored_skill = pd.DataFrame(scored_result, index=scored_index, columns=["ratio",'score','match'])
scored_skill = scored_skill.sort_values(by=["ratio"], ascending=False)

# delete duplicate index
scored_skill=scored_skill[~scored_skill.index.duplicated(keep='first')]
```
Code snippet 5.21 filtering operation

## 5.4 Job Recommendation

Once the user login the website as Applicant, they can upload a CV file in PDF format and preview their CV on the same page based on the VuePdfEmbed library. Once they upload their CV, they can jump into the manage account page and click Job Recommendation on the right-sidebar. The applicant ID and all posted jobs will be transferred to the backend with CVMatchJobAPI. The API will go through the list of job IDs passed from the frontend and get the job description and get the CV path from Firebase. Once it gets the CV file and job description lists, it will be fed into the Advanced_CVMatchJob function to get a list of ranking job recommendation lists. In the Advanced_CVMatchJob function, the scored list of job ranking will be calculated with the ranking algorithm. The function will return the matching ratio with ranking jobs in the last step. The ranking job list will be shown on the website based on their matching score. The first item must be the highly related job with Applicant's CV. Therefore, Applicants could have a clearer overview of high-ranking jobs with their CVs.

## 5.5 Data Visualization

The job dashboard will show three charts, including WordCloud for the job description, scored skill overview and scored applicants ranking overview, which is shown in Figure 5.22. For WordCloud generation, it will pass the job id to the backend with WordCloudAPI. The WordCloudAPI will obtain a job description based on the job id from Firebase collection 'Job'. The obtained job description will be thrown into WordCloud() generate function, and the generated WordCloud file will be uploaded to the Firebase. Once it uploads successfully, it can obtain a WordCloud file from Firebase with a download URL from Firebase Storage, which can be shown on the job dashboard page. For scored skill overview, it will pass the job id to the backend with Job2SkillAPI. The Job2SkillAPI will also obtain the job description and extract skills based on Bert. TfidfVectorizer will obtain the scored list of keywords with TFIDF score. However, the filtered list of skills can be obtained by matching the keywords in two lists. Therefore, a list of scored skills will be returned to the frontend. With the list of scored skills, the skills name will be set to the x-axis label, and the skills score will be set to the y-axis value based on the Echart library. For scored applicants ranking charts, it will get job id and a list of job applicants id to the backend with JobMatchCVAPI. Even though the returned value is the same as the CV ranking, the applicant's name will be set on the x-axis label, and the scores will be set to the y-axis value based on Echart.

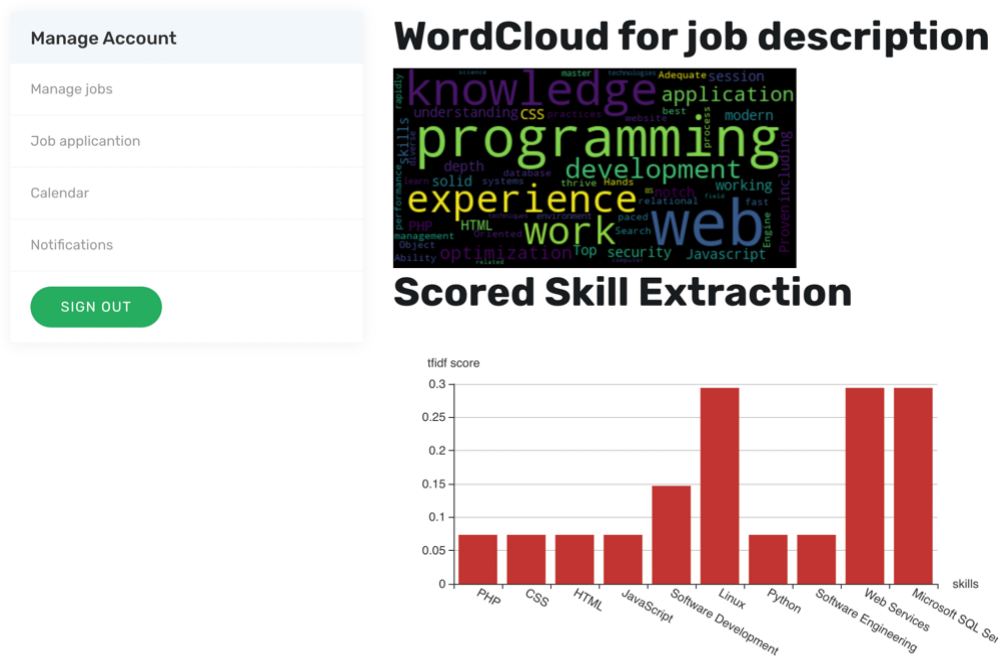

Figure 5.22 Job Dashboard Page for Hunt

## 6. Evaluation and Testing

### 6.1 Usability Testing

This is usability testing will be experimented based on Jakob Nielsen's 10 principles for website interaction and the severity will be classified into following categories [28]:

• 0: Not a usability problem at all

• 1: A cosmetic problem

• 2: A minor problem

• 3: A major problem

• 4: A usability catastrophe

**Visibility of system status**: It will be marked as severity 1. Users can receive a pop-up notification of registration status and know about the error text with registration failure, shown in Figure 6.1. However, the user cannot recognize which information is wrong, and it is a lack of successful notification before changing to the sign-in page.

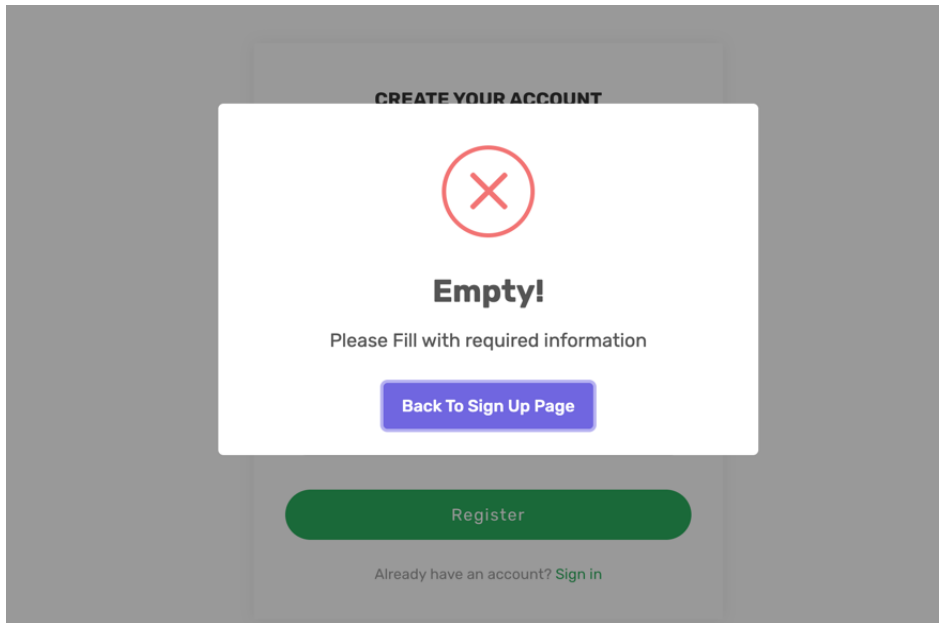

Figure 6.1 Successful Example for Visibility of system status

**Match between system and the real world**: It will be marked as severity 4. If the user does not log in and click on the saved job button or apply job button, the user can understand they do not sign in and go back to the sign-in page when they click 'Back to Sign in Page', which is shown in Figure 6.2. However, it ignores the operation of the hunt, as the hunt has no right to save a job or apply job, which is shown on the hunt page as well.

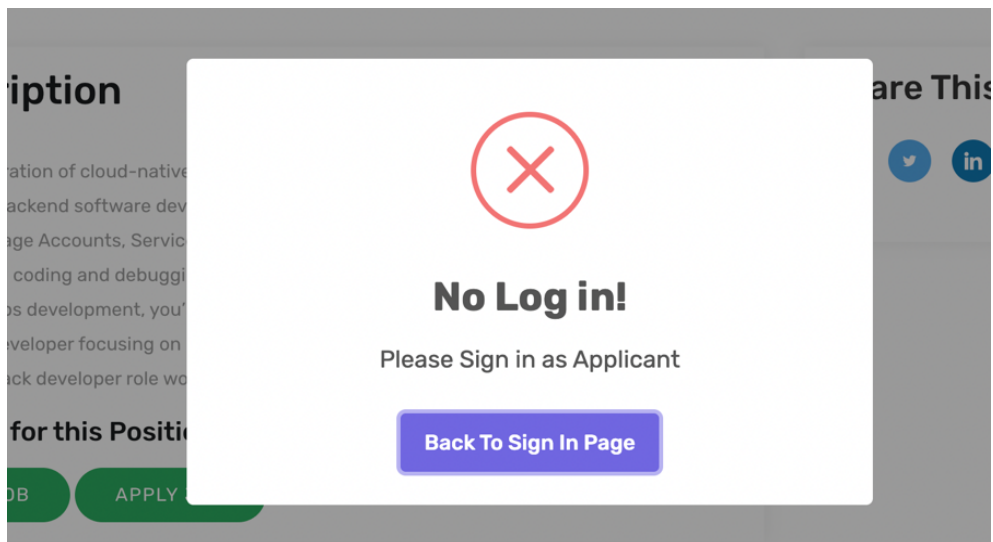

Figure 6.2 Successful Example for Match between system and the real world

**User control and freedom**: It will be marked as severity 3. On the Calendar Page, users can check today's events in the current month or other events in the next four weeks and the previous four weeks, as shown in Figure 6.3. Once they click on the last button, they can still go back to today. Hunt can pick any date and time they prefer to arrange an interview. If they would like to stop the process, they can cancel it with the close tag on the top right corner. However, the hunt has no choice but to save job information before they post a job. For example, if a hunt has some information that is not confirmed, they cannot have a place and status to store their posted job draft.

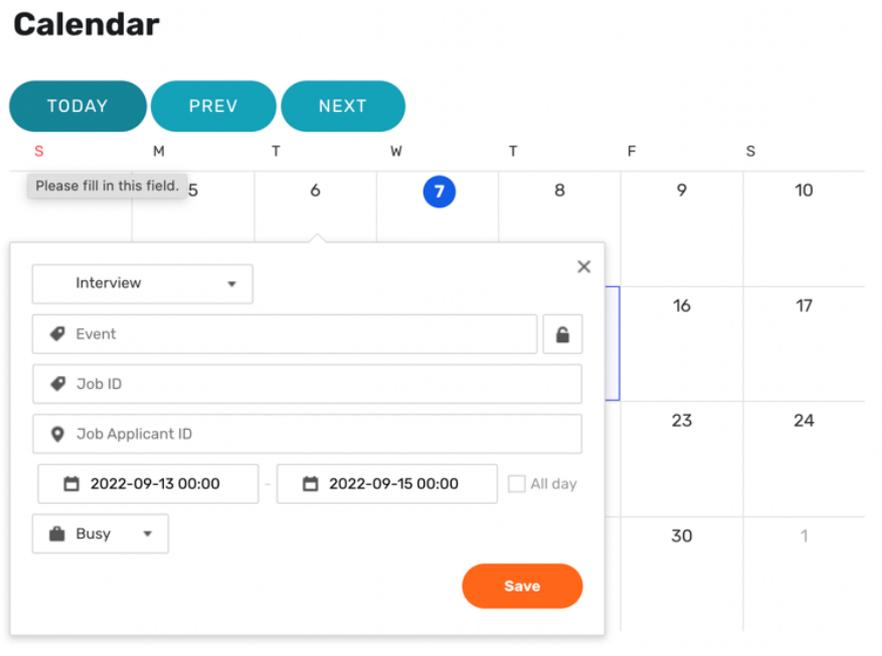

Figure 6.3 Successful Example for User control and freedom

**Consistency and standards**: It will be marked as severity 1. Because the user interface was borrowed from the existing typography remains comparatively the same standard. For example, the fonts on the navigation bar remain the capital at the first letter for each word, as shown in Figure 6.4. Or, in every button, all characters inside were capital. However, some buttons gradually become green from blank while the user hovers the curser on it, whereas some buttons turn their colour into a slighter mode. Besides, when working on a page, the locations of each job are not aligned.

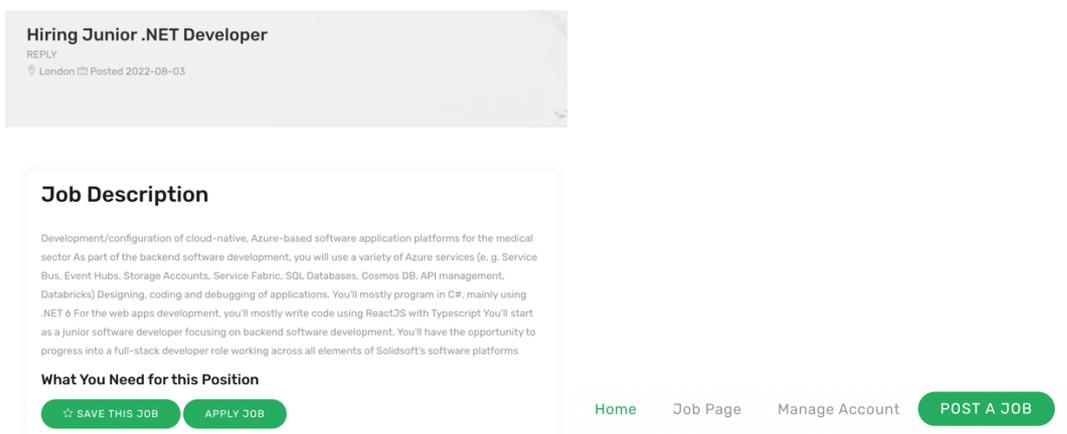

Figure 6.4 Successful Example for Consistency and Standards

**Error prevention**: It will be marked as severity 3. If the hunt accidentally clicks on the delete button to remove the job, they still have the opportunity to cancel their operation shown in Figure 6.5. Besides, it also provides a way for the hunt to confirm their operation. However, suppose the hunt accidentally arranges an incorrect application or inputs the wrong applicant id. In that case, there is no error to pop up, which might have a negative impact on the application process.

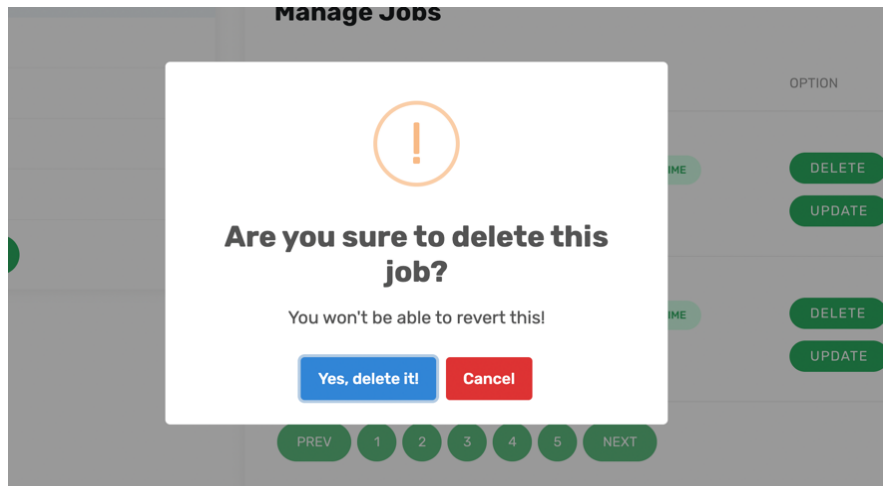
Figure 6.5 Successful Example for Error Prevention

**Recognition rather than recall**: It will be marked as severity 3. Most of the icon is visible on the page, and there is no need to remember the detailed information about the operation. However, while entering the sub-page, it probably has something related to the job which the user should recall. For example, when hunt would like to check applicants by browsing the applicant page, they should enter it via the job application page. But they might forget the job they select, which is the same as in Figure 6.6. Besides, the job dashboard page does not show any information related to the job, except skill extraction and job applicants. Therefore, the hunt might forget which job they prefer to check the result.

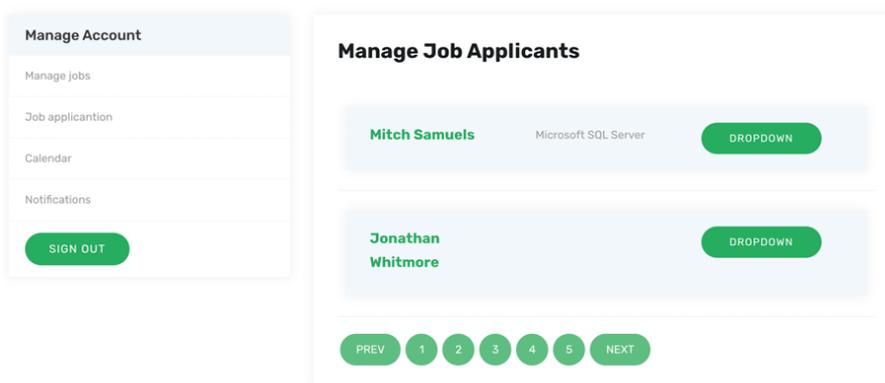
Figure 6.6 Violent Example for Recognition rather than recall

**Flexibility and efficiency of use**: As the current project is a demo for demonstration for the smart job hunter website, it is based on the existing user interface, and some functions have been implemented on my own. Therefore, some advanced features, such as shortcuts for the flexibility of users, have not been considered in this project. However, all the provided functions and API can be applied functionally.

**Aesthetic and minimalist design**: It will be marked as severity 1. Even though the user interface is provided online, some extra features have been removed in this project. For example, the integrated blog has been removed from the original website as the blog page does not stand for the major component. The major parts in this demo are the job page, job application and other operations related to the job application, which are essential. However, there are still some unnecessary icons and links at the bottom of the footer, and most of them are not useful to this project, which is shown in Figure 6.7.

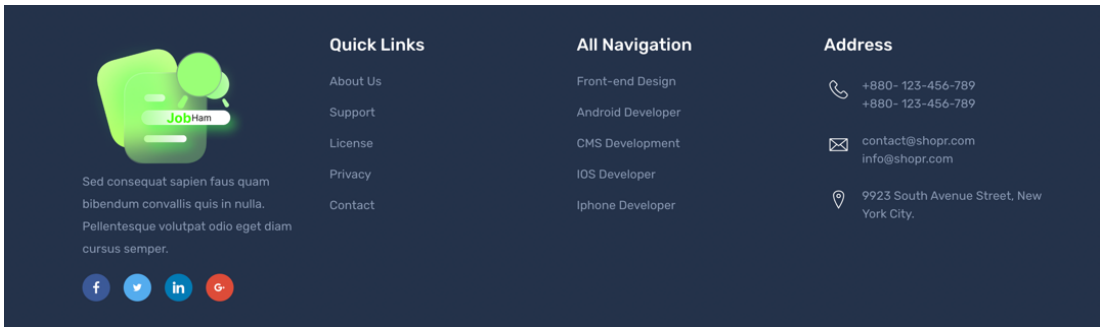

Figure 6.7 Violent Example for Aesthetic and minimalist design

**Help users recognize, diagnose, and recover from errors**: It will be marked as severity 2. As the hunt has the same preview as an applicant, they still can see buttons which are not related to their operation, such as 'apply job' and 'saved job'. But they cannot continue their process if they click the above buttons, and it will show an error message to notify them they are not acting as the applicant, and they cannot continue this operation, as shown in Figure 6.8. However, it does not provide a solution for the hunt as it should not show the above buttons on the pages, and it aims to stop hunt from continuing its wrong operation.

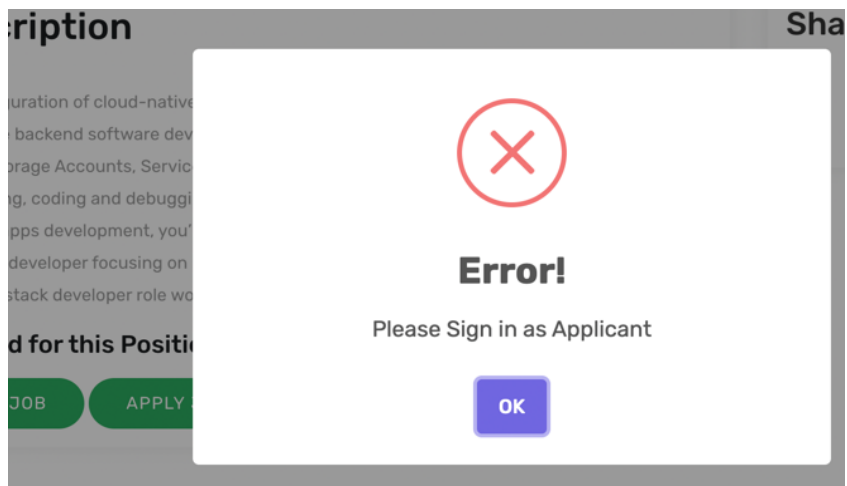

Figure 6.8 Successful Example for Helping users recognize, diagnose and recover from errors

**Help and documentation**: The website will not need additional documents to provide as it is the process of job application, and all related roles follow the real-world logic, such as the admin can delete accounts, the hunt can manage jobs they posted, and the applicant can apply for their jobs. Besides, all the operations are different in the sub-page of the manage account page, and it is clear for every role to operate. Therefore, no extra help would be provided on this website.

## 6.2 API Testing

The backend API provides a way to connect the backend and frontend. Based on the API, frontend and backend can communicate with their preferred data. The endpoints are classified with their function. For example, the endpoint for JobMatchCVAPI is MatchCV. Table 6.1 show that all backend API is applied in this project. To test API in this project, Postman is a way to evaluate APIs. In the test process, the URL should be input into the request URL, and input data should be input into the parameter and change the method to GET. Once it sends the request by API, the result of it will be shown on the website.

| API Name | URL | Input Data |
|---|---|---|
| **JobMatchCVAPI** | LocalHost/JobMatchCV/<Job>/<Applicants> | Job id, list of applicant id |
| **CVMatchJobAPI** | LocalHost /CVMatchJob/<ApplicantID>/<Job> | Applicant id, list of job id |
| **WordCloudAPI** | LocalHost /WordCloud/<JobID> | Job id |
| **Job2SkillAPI** | LocalHost /Job2Skill/<JobID> | Job id |

Table 6.1 the list of Backend API

For JobMatchCV API, the input data job id and list of applicant id would be input in the URL field. The return data will be in JSON format and shown on the screen in Figure 6.1. The result will save the score and match_list for every applicant as JSON.

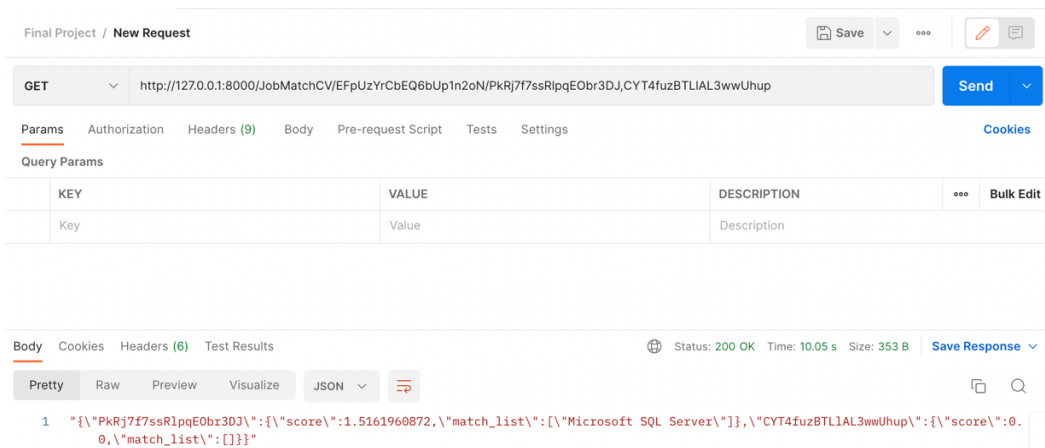

Figure 6.1 JobMatchCV API testing Example

For CVMatchJob API, the input data applicant id and list of job id would be input in the URL field. The return data will be in JSON format and shown on the screen in Figure 6.2. As a result, every job's score will be saved as JSON.

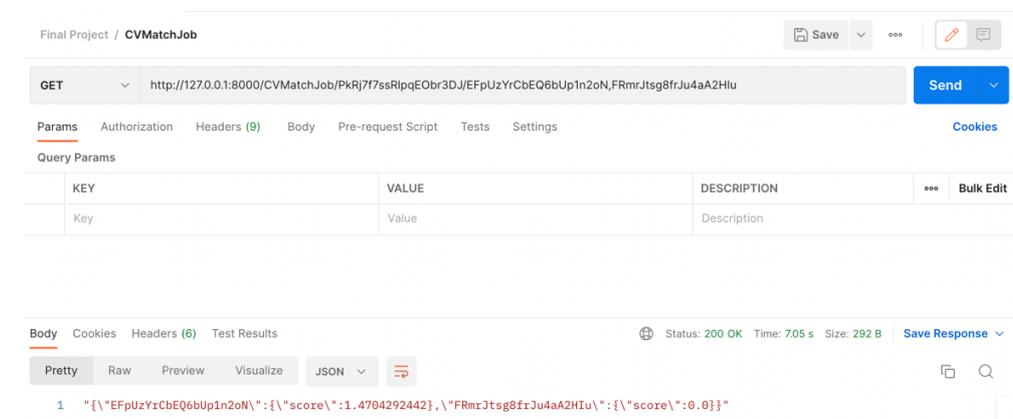

Figure 6.2 CVMatchJob API testing Example

For WordCloud API, the input data job id would be input in the URL field. The return data will be in JSON format and shown on the screen in Figure 6.3. As a result, the job id will be saved as JSON.

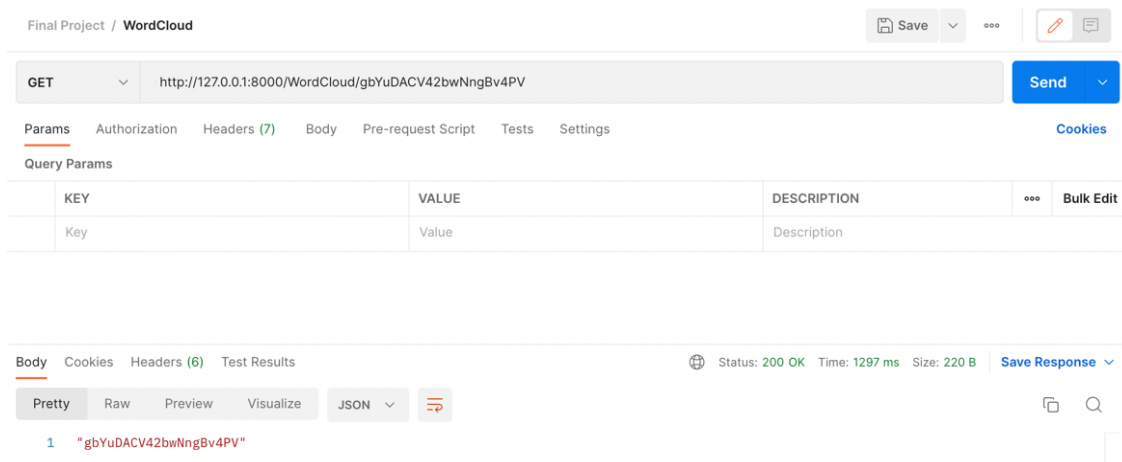

Figure 6.3 WordCloud API testing Example

For Job2Skill API, the input data job id would be input in the URL field. The return data will be in JSON format and shown on the screen in Figure 6.4. In the result, the list of skills, its score and match with skills extracted from TFIDF tokenization will be saved as JSON.

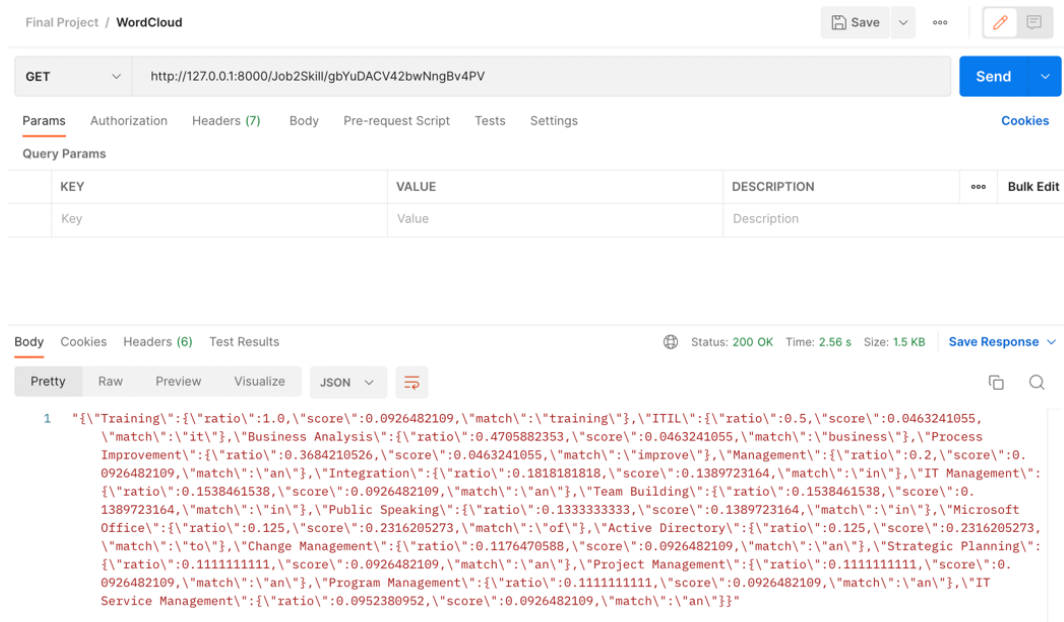

Figure 6.4 Job2Skill API testing Example

## 6.3 Automatic Keyword Extraction Model Evaluation

In this section, the evaluation matrixes for CV2Skill and Job2Skill model have been provided with a reasonable result.

### 6.3.1 CV2Skill Model Evaluation

For the CV2Skill model, three matrixes exist to evaluate the model, including accuracy, precision, recall and F1 score, which is related to TruePositive, TrueNegative, FalsePositive and FalseNegative. TruePositive means that the model correctly predicts positive class, and TrueNegative means that the model correctly predicts negative class. Besides, FalseNegative implies that the model incorrectly predicts a negative class and FalsePositive means that the model incorrectly predicts a positive class. Accuracy will calculate the proportion of correct predictions in the total number of all predictions for the test data, which will be shown below [29]. Precision will calculate the proportion of the sum of true positives in the total numbers, which will be shown below [30]. True positive means that the number that models correctly predicts the outcome, while false positive means that the number that models incorrectly predict the outcome. The recall will compute the proportion of the correct prediction in the total number of positive forecasts for the sample data, which will be shown below. The number of positive prediction outcomes means that the total number of results should be positive. F1 score is based on Recall and Precision by calculating double of the average of Recall and Precision, which will be shown below [31].

$$Accuracy = \frac{TruePositive + TrueNegative}{TruePositive + TrueNegative + FalsePositive + FalseNegative}$$

$$Precision = \frac{TruePositive}{TruePositive + FalsePositive}$$

$$Recall = \frac{TruePositive}{TruePositive + FalseNegative}$$

$$F1Score = 2 \times \frac{Recall \times Precision}{Recall + Precision}$$

As the model trained ten epochs on the AutoDL [27] platform, each email address, Name and skill label matrix will be shown below. In 10 epochs, the entity of email address and Name show an increasing trend, whereas skills show an oscillating trend overall for Precision in Figure 6.5, which means that the name and email address has a higher probability of classification precision compared with entity skills labels. Regarding recall, the entities of skills and Name climb around 85% while the entity of email address remains around 80% in Figure 6.6, which means that the important entities can be extracted most of the time. For F1Score shown in Figure 6.7, the entity of email address is around 80%, while the entity of Name and skills show the same trend as it shows in the recall matrix, which means that the classification mode is stable to identify Name, email address and skills entitles.

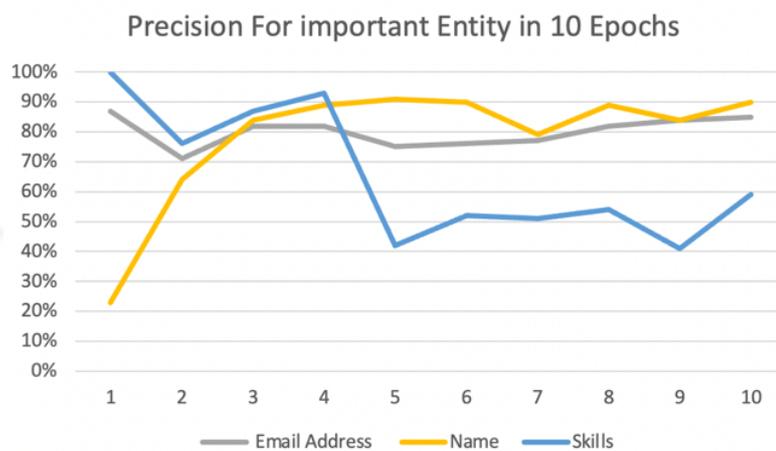

Figure 6.5 Precision for important entity

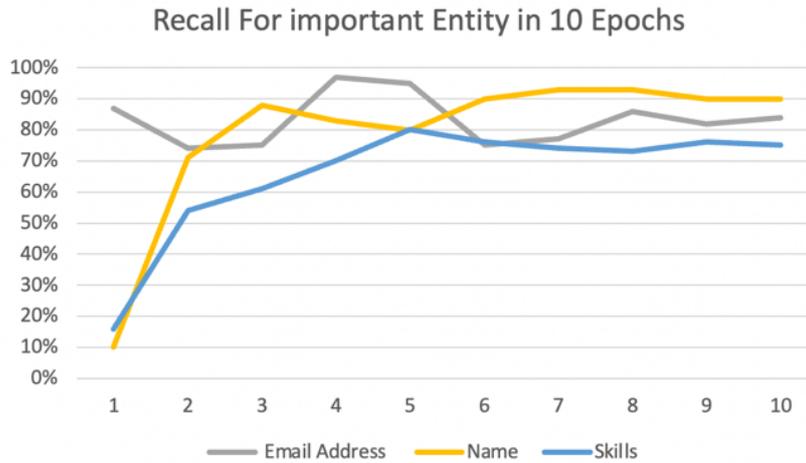

Figure 6.6 Recall for important entity

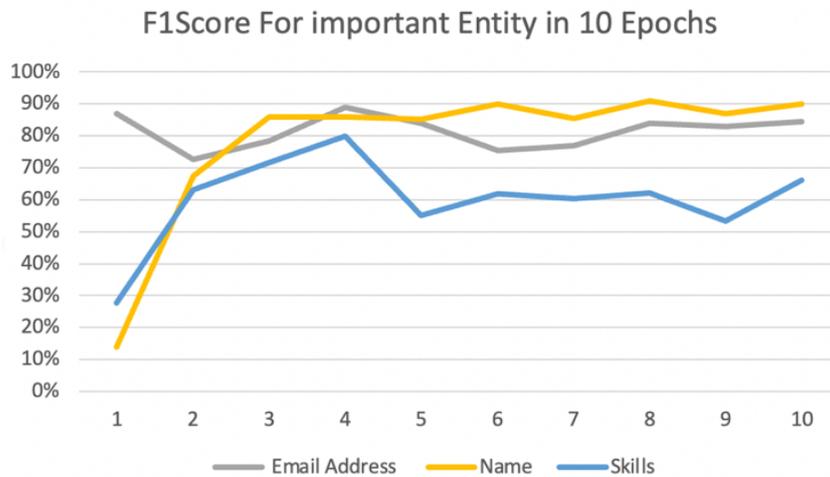

Figure 6.7 F1Score for important entity

Besides, except for model accuracy performed on the test data, the loss can be another matrix to evaluate the whole model, which is CrossEntropyLoss to compute the loss of classification problem. CrossEntropyLoss aims to calculate the distance between prediction and labels by computing LogSoftmax() and NLLLoss(), which is very useful in the classification task. LogSoftmax is computed by the SoftMax function log, while NLLLoss stands for Negative Log-Likelihood Loss. If CrossEntropyLoss has a stable decreasing trend, the CV2Skill model is good for identity entities in a sentence, as shown in Figure 6.8. It shows that the loss has a steady decreasing trend for the training dataset and validation dataset while the accuracy for both datasets, which means that the CV2Skill model gets a higher accuracy with almost 90% for the validation dataset. Compared with other models, such as BERT-LMCRF, even though it still has lower accuracy, it still has a reasonable result.

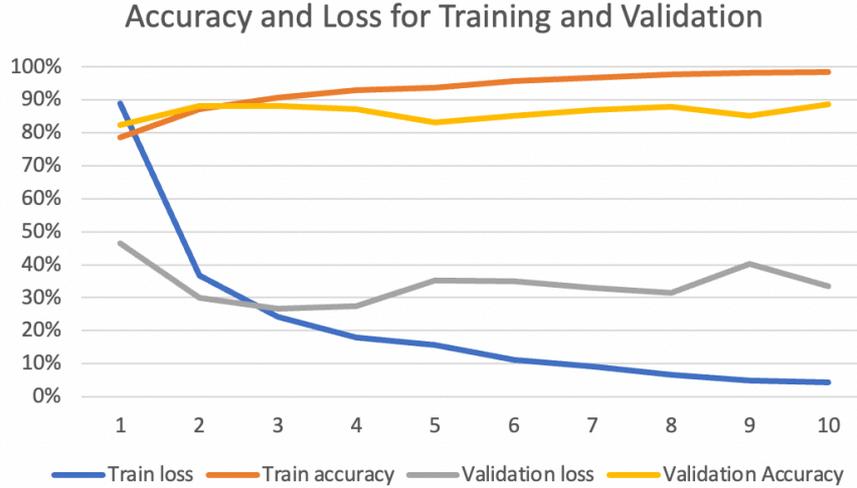

Figure 6.8 Accuracy and Loss for Training and Validation

### 6.3.2 Job2Skill Model Evaluation

For Job2Skill model, there exists three matrixes to evaluate the model, which contains Mean Reciprocal Rank (MRR) [32], Normalized Discounted Cumulative Gain (NDCG) [33] and Recall. Mean Reciprocal Rank is a matrix to evaluate the ranking item with the query item, which is shown in Formular. For example, for a single query, if the query is cat, the returned value 'cat' ranked the fifth in the list. Therefore, the MRR for query 'cat' is 1/5. Before mention the Normalized Discounted Cumulative Gain, Discounted Cumulative Gain should be explained firstly. Discounted Cumulative Gain aims to measure the quality of ranking while cumulative gain is the score of relevance ratio with query. And the DCG will be provided by dividing $log_2(i+1)$ with cumulative gain, which i represents the $i_{th}$ items. Besides, Ideal Discounted Cumulative Gain (IDCG) will be provided by ranking the result items in its ideal situation, which will be shown in Formular. Thus, NDCG will be provided by dividing DCG with IDCG. Recall is the same as the before mentioned in the CV2Skill model evaluation.

$$MRR = \frac{1}{Q}\sum_{i=1}^{Q}\frac{1}{rank_i}$$

$$DCG = \sum_{i=1}^{Q}\frac{relevance_i}{log_2(i+1)}$$

$$NDCG = \frac{DCG_n}{IDCG_n}$$

As the model has trained 20 epochs on the third-party platform, AutoDL, the MRR, variant NDCG and variant Recall have shown the trend in the following Figures. MRR is steadily increasing and up to 94%, which is higher than other models shown in Table 6.2. NDCG@100 means the NDCG score within the top 100 skills, while recall@100 means the Recall score of the top 100 skills. While recall@100 shows a higher probability than Recall@50, the score of NDCG@100 is almost the same as the score of NDCG@50 shown in Figure 6.9 and Figure 6.10. Therefore, regarding the score of Recall and MRR, it is suitable to identify over 50 skills in a sentence, which shows a reasonable result.

| Model Name | MRR |
|---|---|
| **BiLSTM [34]** | 0.8565 |
| **BiGru [35]** | 0.8716 |
| **BERT-XMLC [36]** | 0.9019 |
| **Job2Skill** | **0.9382** |

Table 6.2 Mean Reciprocal Rank Comparison

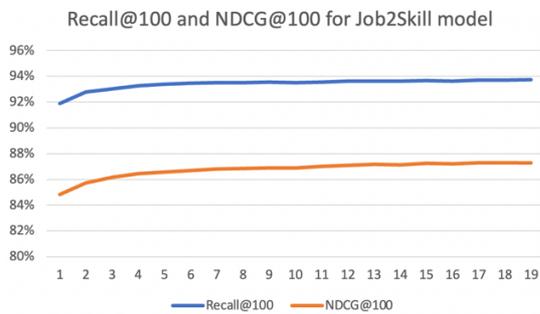

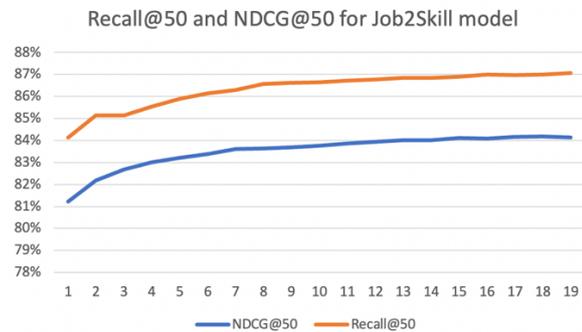

Figure 6.9 Recall@100 and NDCG@100 score    Figure 6.10 Recall@50 and NDCG@50 score

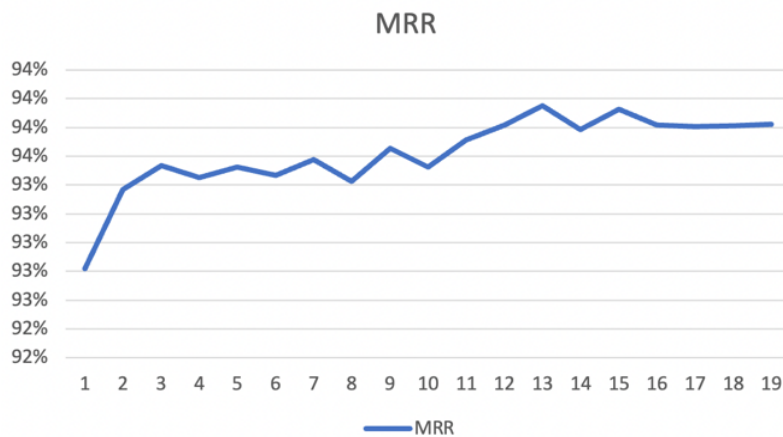

Figure 6.11 Mean Reciprocal Rank score

Besides, except for the above matrix, loss for validation dataset is another matrix to evaluate model, BCEWithLogLoss to compute loss for the multi-classification task. As the last layer in the model does not contain an activation layer, the activation function will be added to the loss function. Therefore, BCEWithLogLoss is a loss function to prevent the sigmoid layer and BCELoss at the last stage. BCELoss is a categorical cross-entropy loss function for multi-classification tasks. The validation loss has been shown in Figure 6.12. It offers an overview of decreasing trend for validation loss, which means that it still has improvement for further training.

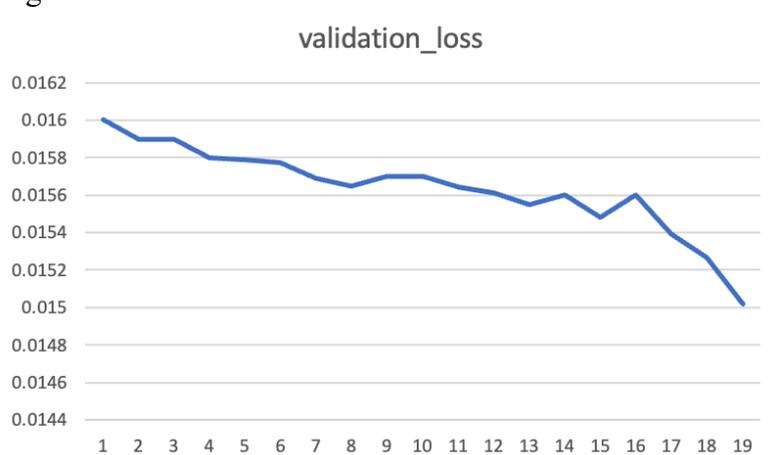

Figure 6.12 validation Loss while model training

## *7. Conclusion and Future Work*

The project aims to conduct a website to help graduates find their job and head hunter to hire high-related employees more effectively. All the requirements mentioned in this project have been implemented, including the job recommendation and CV ranking implementation, even with the optional requirements. Even though the architecture of the Job2Skill and CV2Skill model is essential, it provides a reasonable result for this project and can extract a keyword from the job description and CV file. Based on the keyword extraction, the website can automatically provide a ranking list of job recommendations and applicants. Moreover, job/CV ranking algorithms will be combined with keyword extraction, which could provide a list of ranking jobs and candidates. The job dashboard offers a way to have an overview of the job description required skills and marked job applicants. Besides, an integrated calendar helps the user to improve their user experience while checking the important date of the job application and interview. And hunt can arrange an interview with an integrated calendar and send a notification to job applicants.

The CV2Skill and Job2Skill model still have low accuracy for some careers as the datasets for model training does not have enough data for specific careers. Therefore, posted job datasets could be provided by clawing datasets on Indeed or other job-hunting websites and collect posted jobs for each career might contain enough data. Another reason for lower performance on identification might come from the model design, especially for the CV2Skill model. As the CV2Skill model train on the basic model and the architecture of it has no change, the model cannot classify the entity in a higher performance, especially for entity designation. Also, due to ethical issues, few datasets are related to the resume dataset. Therefore, the best step to collecting high-quality resume datasets is to cooperate with related cooperation and company and ask for help from them. The CV/Job ranking algorithm considers the occurrence ratio in the whole document instead of the skill vocabulary list. If possible, it will be revised to evaluate the occurrence ratio in the list of skill vocabulary.

In short, the requirements have been successfully implemented and integrated into the website. The natural language processing model has been applied in this project and provides a reasonable result for users. Users can process their job applications more effectively, which is the goal of the project and has been achieved.

# Appendix

**Appendix 1**: The updated Gantt chart has been shown as below based on the project process.

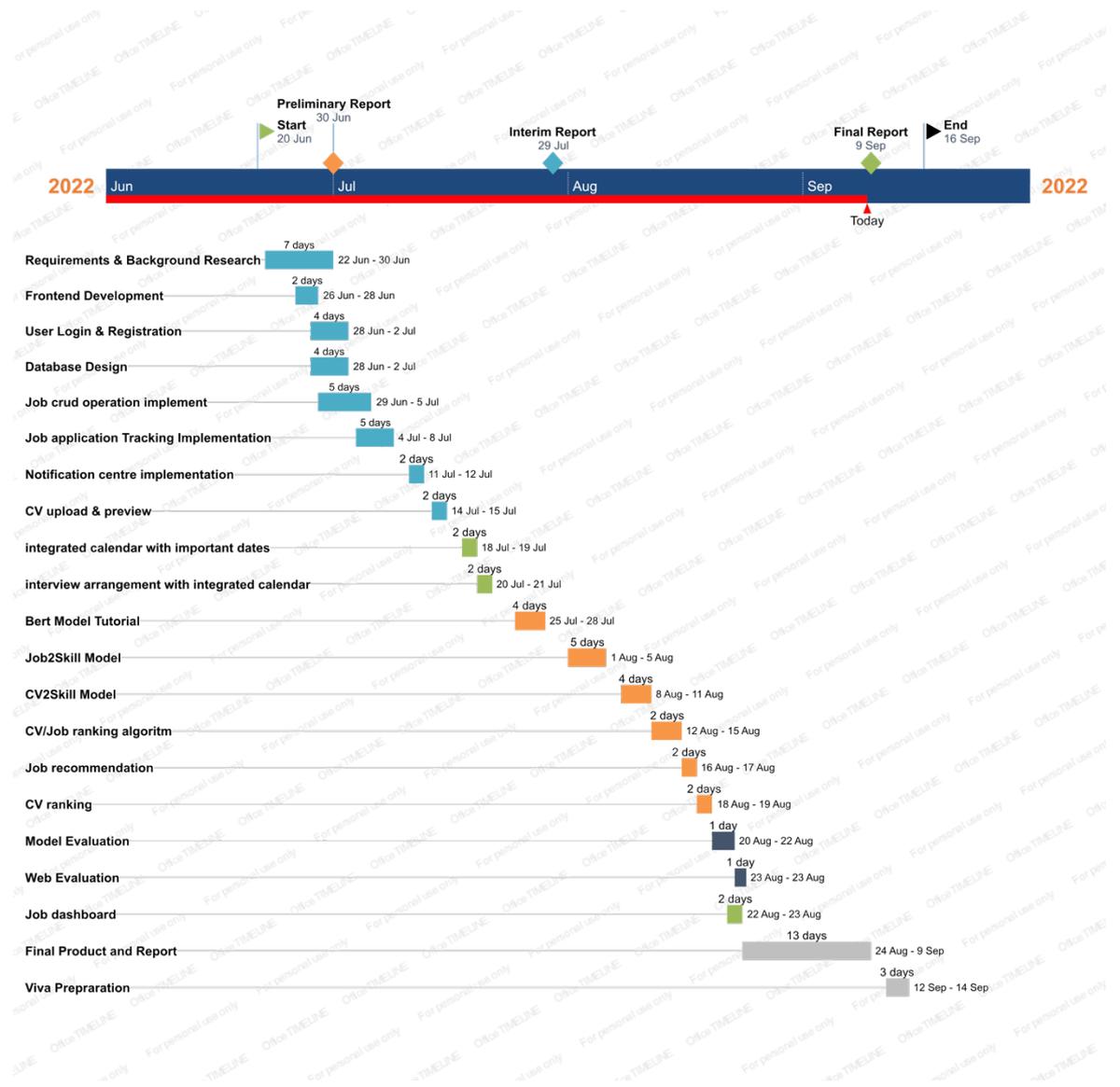